\documentclass[runningheads]{llncs}
\usepackage{graphicx}
\usepackage[ruled,vlined]{algorithm2e}
\begin{document}
\title{Hybrid Supervision Learning for Pathology Whole Slide Image Classification}


%
%

\author{Jiahui Li \inst{1} \and
Wen Chen\inst{1} \and
Xiaodi Huang \inst{1} \and
Shuang Yang \inst{1} \and
Zhiqiang Hu \inst{1} \and
Qi Duan \inst{1} \and
Dimitris N. Metaxas \inst{2} \and
Hongsheng Li \inst{3,4} \and
Shaoting Zhang\inst{1,4}\thanks{This study has been financially supported by fund of Science and Technology Commission Shanghai Municipality (19511121400), also partially supported by the Centre for Perceptual and Interactive Intelligence (CPII) Ltd under the Innovation and Technology Fund. Code of this paper is available at https://github.com/JarveeLee/HybridSupervisionLearning\_Pathology} }
\authorrunning{Jiahui Li. et al.}
%
\institute{SenseTime Research, Shanghai, China \\
\email{\{lijiahui,chenwen,huangxiaodi,yangshuang1,huzhiqiang,duanqi,zhangshaoting\}@sensetime.com}\\
\and
Rutgers University, New Jersey, USA\\
\email{dnm@cs.rutgers.edu}\\
\and
The Chinese University of Hong Kong, Hong Kong, China\\
\email{hsli@ee.cuhk.edu.hk}\\
\and 
Centre for Perceptual and Interactive Intelligence (CPII) Ltd, Hong Kong, China\\
}
\maketitle              
\begin{abstract}
Weak supervision learning on classification labels has demonstrated high performance in various tasks, while a few pixel-level fine annotations are also affordable. Naturally a question comes to us that whether the combination of pixel-level (e.g., segmentation) and image level (e.g., classification) annotation can introduce further improvement. However in computational pathology this is a difficult task for this reason: High resolution of whole slide images makes it difficult to do end-to-end classification model training, which is challenging to research of weak or hybrid supervision learning in the past. To handle this problem, we propose a hybrid supervision learning framework for this kind of high resolution images with sufficient image-level coarse annotations and a few pixel-level fine labels. This framework, when applied in training patch model, can carefully make use of coarse image-level labels to refine generated pixel-level pseudo labels. Complete strategy is proposed to suppress pixel-level false positives and false negatives.  A large hybrid annotated dataset is used to evaluate the effectiveness of hybrid supervision learning. By extracting pixel-level pseudo labels in initially image-level labeled samples,  we achieve 5.2\% higher specificity than purely training on existing labels while retaining 100\% sensitivity, in the task of image-level classification to be positive or negative. 

\end{abstract}
\section{Introduction}
Hybrid supervision learning on various levels of annotations has shown its effectiveness in various machine learning applications \cite{mlynarski2019deep,thoracic,huang2020rectifying} . However, we find those tasks in computational pathology is more challenging as high resolution ( over 100,000 $\times$ 100,000 pixels ) of whole slide images makes it difficult to conduct end-to-end training of deep learning models, which is challenging to perform weak or hybrid supervision learning research in the past \cite{mlynarski2019deep,thoracic,pei2019multimodal,heng2018hybrid,robert2018hybridnet,he2009learning,shah2018ms}. 

To handle aforementioned challenges, a novel hybrid supervision learning framework is proposed for whole slide images classification to distinguish positive or negative. Due to high resolution we can only train deep learning model on patches, and propose a well-designed strategy to modify pixel-level pseudo labels on patches, according to image-level labels: A positive image is guaranteed to contain at least one positive patch, while a negative image shall be entirely pixel-wisely negative. Secondly pixel-level errors, false negatives and false positives, are gathered during pseudo labels generation. We perform re-weighting on pixel-level pseudo labels of patches from positive images, converting false negatives to true positives and false positives. We use hard negatives patches from negative images with a higher sampling ratio to train model, to further convert false positives to true negatives. Because noise tolerant deep learning model can discriminate a pattern as negative if during training it is mostly labeled as negative. As for the few pixel-level annotations, we use them to initialize pre-trained model and to mix a constant ratio in each training batch with pixel-level pseudo labels to regularize training. With such strategy, without end-to-end training, we can make best use of image-level labels, pixel-level fine-grained labels and pseudo labels on unlabeled areas. 

The main contributions of our hybrid supervision learning framework is to utilize both the limited amount of pixel-level annotations and the large number of image-level labels. Without end-to-end image-level training, image-level labels are used to refine pseudo labels on the entire training dataset, with well-designed strategy to control false positives and false negatives. We evaluated the framework on a large whole slide images dataset of histopathology. According to experiments results, the hybrid supervision learning shows specificity 8.92\% better than image-level training and 5.2\% better than pixel-level training, while retaining 100\% sensitivity. 

\begin{figure}[!t]
\centerline{\includegraphics[width=0.7\columnwidth]{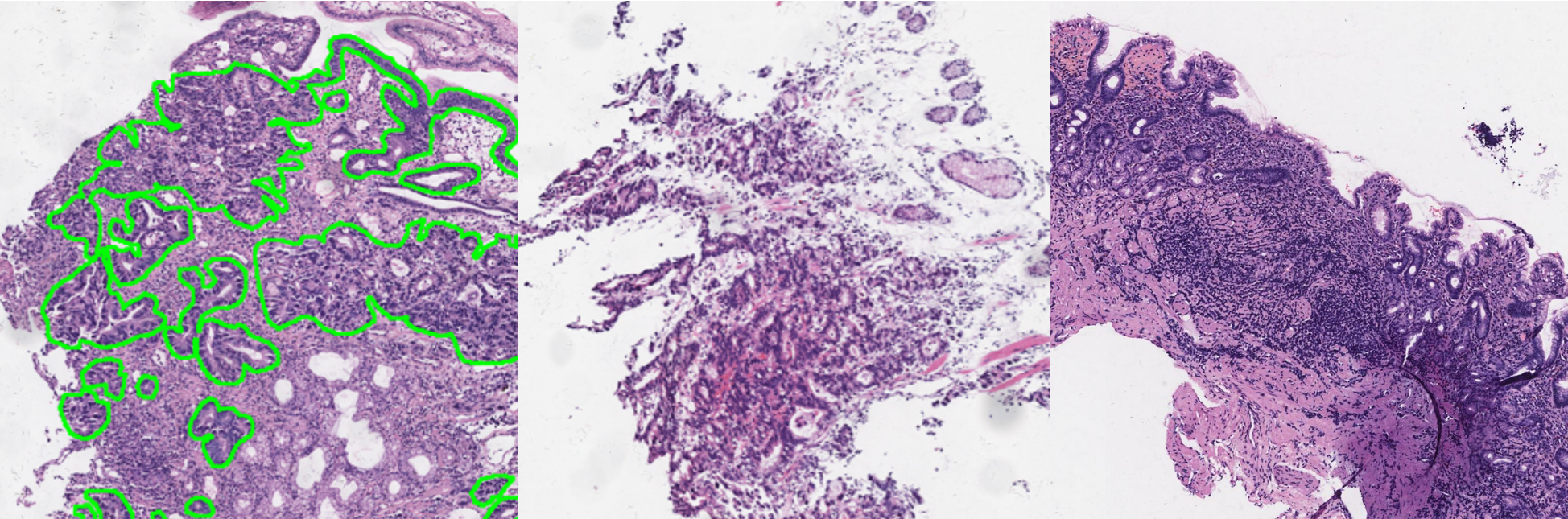}}
\caption{Illustration of hybrid supervision data, containing 2 types of label. The first expensive and rare type is pixel-level fine-grained labels, contoured by green lines. The second type is image-level labeled images. The rest two images are image-level positive and negative.}
\label{mixsupervision}
\end{figure}

\section{Related work}
For hybrid supervision learning: In most medical image tasks, hybrid supervision is a common data situation, with large amount of clinical reports and small group of segmentation annotations shown in Fig. \ref{mixsupervision}. However focus was mostly given to semi and weak supervision learning \cite{SemiBox,SemiSeg3,Semi5Selftrain2,signet,xie2019self}, defined on single format then to make use of unlabeled images. There exists some achievements related to our framework\cite{mlynarski2019deep,thoracic,huang2020rectifying}. Our framework is different from theirs because end-to-end classification training with entire image input is difficult in whole slide images. Instead only patches are involved in our total pipeline, image-level labels are not directly related to loss calculation. As for computational pathology: Usually the size of one whole slide image is 100,000 $\times$ 100,000 pixels, which is too big to run directly on GPU, therefore most of previous contributions process images in a two-stage manner \cite{shaban2019context,hou2016patch,postrate,nagpal2019development,nature,takahama2019multi,zhou2019cgc,khened2020generalized}. In \textbf{the first stage}, discriminative patches are extracted from whole slide image by specific patch models, in the second stage a whole slide image classifier is trained by the selected patches from first stage.

\section{Hybrid supervision learning for whole slide image classification}

\begin{figure}[!t]
\centerline{\includegraphics[width=0.9\columnwidth]{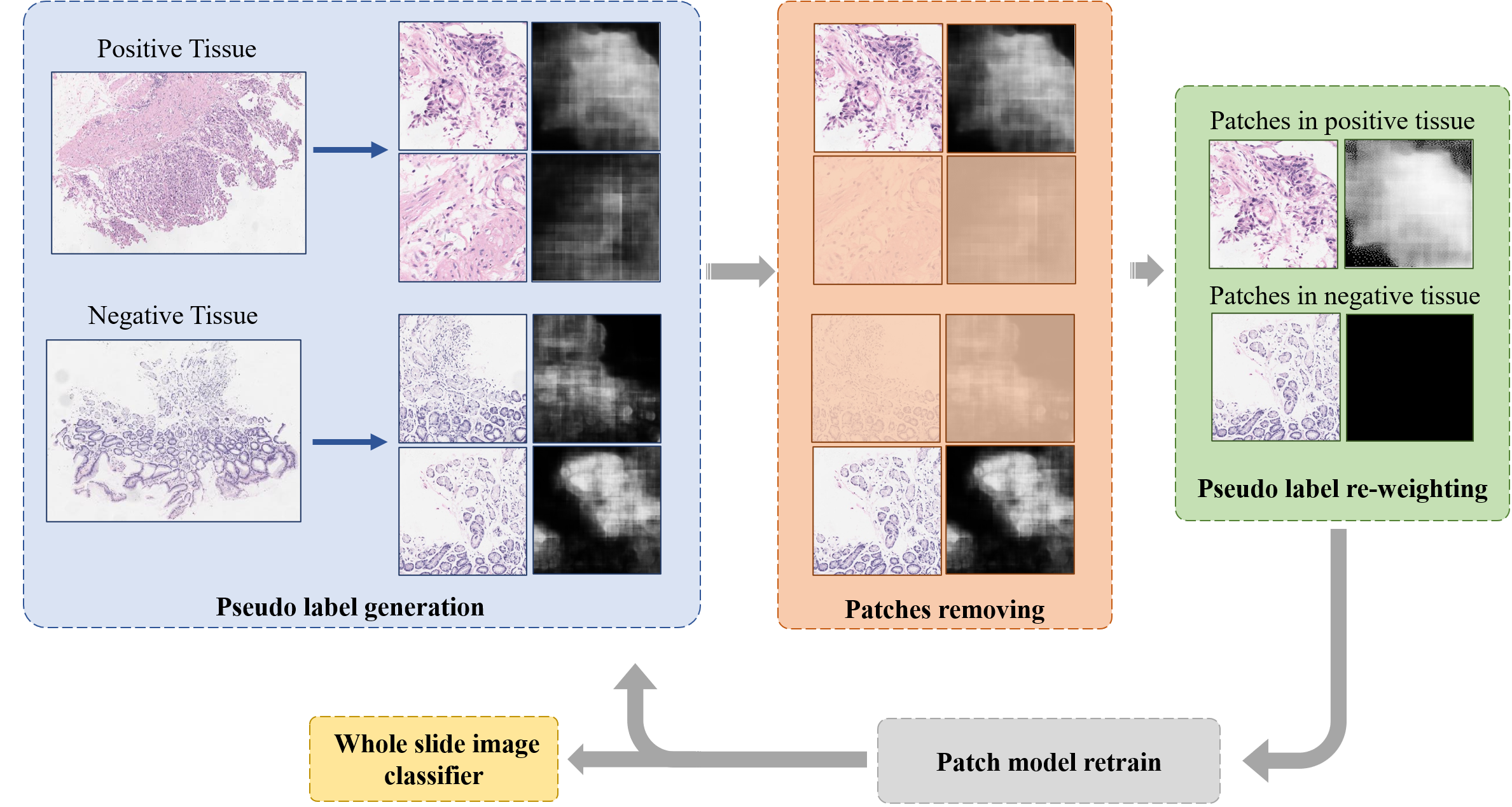}}
\caption{The overall pipeline of our hybrid supervision learning for whole slide image. The image-level label guided pixel-level pseudo label generation is iterated in this manner. The gray scale map is the predicted probability map of patch model.}
\label{majorpipline}
\end{figure}

The main objective of hybrid supervision learning is to use image-level labels to refine pixel-level pseudo labels on patches, without the entire huge image input. Pipeline is shown in Fig. \ref{majorpipline} and Algorithm. 1, which involves two stages: patch segmentation, whole slide image classification.

\begin{algorithm}[!t]
\KwData{

$y_{p}$: pixel-level fine-grained label from pathologists. 

$\hat{y_{p}},\hat{y_{p+}},\hat{y_{p-}}$: pixel-level pseudo label, from positive and negative images, created by models and image-level labels guided re-weighting.  

$y_{i}$: image-level label from clinical reports.  

$\theta_{1}$: patch segmentation model initially trained by $y_{p}$, outputs the probability that each pixel of input patch is positive. 

$\theta_{2}$: whole slide image classifier, outputs the probability that input whole slide image is positive. 

$I$: high resolution whole slide image. 

$x$: image patches. 

$T$: patch removing threshold. 

$V$: re-weighting constant.

$R$: sampling ratio of $y_{p}$, $\hat{y_{p+}}$ and  $\hat{y_{p-}}$ .  

$K$: select top K patches as input to train whole slide image classifier.

}
 \While{the model do not converge}{
   \textbf{Stage 1: patch segmentation} \;
   \textbf{E-step: } $\hat{y_{p}}$ $\leftarrow$ $P(y_{p}\bigl| $x$,\theta_{1})$ \; 
   
  Remove those patch $x$ whose maximum pixel-level $\hat{y_{p}}$ in this patch is less than $T$ \;
  
  \eIf{$y_{i}$ == 1}{
   $\hat{y_{p+}}$ $\leftarrow$ $\hat{y_{p}}$ $\times$ V (pseudo labels)\;
   $\hat{y_{p+}}$ $\leftarrow$ 1.0 if $\hat{y_{p+}}$ \textgreater 1.0 (clip within 1.0, to remain true positive)\;
   $\hat{y_{p+}}$ $\leftarrow$ 0.0 if $\hat{y_{p+}}$ \textless 0.01 (clip to 0.0 if lower than 0.01, to remain true negative which was  slightly scaled up by $V$)\;
   }{
   $\hat{y_{p-}}$ $\leftarrow$ 0 (hard negative labels)\;
  }
  
  \textbf{M-step: } Retrain patch segmentation model $\theta_{1}$ at proper sampling ratio of $R$ in each training batch, and by pixel-level soft label cross entropy loss = -$\frac{1}{N}$   $\sum_{N}$ $\sum_{ y =  y_{p} \cup \hat{y_{p+}} \cup \hat{y_{p-}}}$ $y \times log P(y_{p} \bigl| x,\theta_{1}) +  (1.0 - y) \times log (1.0 - P(y_{p} \bigl| x,\theta_{1}))$ \;
  
  \textbf{Stage 2: whole slide image classification} \;
  
  Select top $K$ patches for each whole slide image according to pixel-level maximum $P(y_{p} \bigl| x,\theta_{1})$\;
  
  $P(y_{i} \bigl| I,\theta_{2})$ = $\frac{1}{K}   \sum_{K}  P(y_{i} \bigl| x,\theta_{2})$
  
    Train classification model $\theta_{2}$ by Loss = -$\frac{1}{N}  \sum_{N} y_{i} \times log P(y_{i} \bigl| I,\theta_{2}) +  (1.0 - y_{i}) \times log (1.0 - P(y_{i} \bigl| I,\theta_{2}))$

  \textbf{Convergence criteria}: For each round of stage1, perform stage2 training. Select the round whose stage2 training loss is the lowest, which means the top K patches selected in that round by stage1 is the optimal to fit image-level annotations.

 }

 \caption{Pipeline of hybrid supervision learning.}
\end{algorithm}

A clinically effective system can tolerate false positives for further recheck by pathologists, while false negatives are vital faults for patients. Thus we need to locate the positive patterns in positive images as much as possible, while precluding more negative images. Our goal is to maintaining 100\% sensitivity and pursue higher specificity to reduce workloads. Under this consideration, comparing positive images and negative images, patterns that only exist in positive images shall all be suspicious and at least one patch is responsible for image-level positive diagnosis. Also, negative patterns that exist in negative images, can certainly be regarded as true negative patterns, and positive predictions in negative images can be regarded as hard negative patches. At the same time, it is the fact that collecting negative images is much easier than positive images. With sufficient suppression of various hard negative patches, we give a positive growing tendency to patterns in positive images. Only the confidence of those true positive patterns, not covered by negative patches, will be able to gradually grow up and reach 1.0 eventually. Those noisy false positive introduced by growing tendency will be suppressed by various hard negative patches and higher sampling ratio during deep learning model training.

As for pixel-level fine-grained labels $y_{p}$, the best performance is certain if we perform all the pixel-level fine-grained labels for all the positive images, however with limited budget we could just afford a few, whose quantity is significantly smaller than the image-level labels. To involve pixel-level fine-grained labels $y_{p}$ into training, firstly it is used to initialize a pre-trained model $\theta_{1}$, to minimize false positives and false negatives at the beginning. Secondly it consists of a large ratio in each training batch. 

To implement our intuition,  shown in Fig. \ref{majorpipline}, we separate large size of whole slide images into patches, then develop an \textit{Expectation-Maximization (EM)}-like method to make full use of three types of annotation: image-level labels $y_{i}$, pixel-level fine-grained labels $y_{p}$ and pixel-level pseudo labels $\hat{y_{p}}$. In the E-step, pixel-level pseudo labels $\hat{y_{p}}$ are firstly created from segmentation confidence map of patches from both positive and negative images. We remove all the patches from both positive and negative images whose maximum pixel-level positive confidence is less than a threshold $T$, to reduce the number of training patches. Provided with image-level labels, we obtain hard negative patches $\hat{y_{p-}}$ from negative images and noisy pseudo labeled positive patches from positive images. Then we multiply a weight $V$ ($V>1$) on noisy pseudo labels in positive patches and clipped withing 1.0, assigning them as $\hat{y_{p+}}$, which transforms false negatives to true positives and false positives, while keeping true positives as the same.  In the M-step, patch segmentation model is then trained on a sampling ratio of pixel-level fine-grained labels $y_{p}$, pseudo labeled positive patches $\hat{y_{p+}}$ and hard negative patches $\hat{y_{p-}}$. For sampling ratio of $y_{p}$, $\hat{y_{p+}}$ and  $\hat{y_{p-}}$ in each training batch, hard negative patches $\hat{y_{p-}}$ shall be much more than pseudo labeled positive patches $\hat{y_{p+}}$ so that if one pattern is both labeled as negative in hard negative patches and positive in pseudo labeled patches, model can still regard such pattern as negative for much higher sampling ratio, which transforms false positives to true negatives. Only those patterns not suppressed by negative patches are possible to be eventually discriminated as positive by models. Such procedure is iterated for several rounds and we evaluate sensitivity and specificity for each to decide when to stop. Loss function in patch segmentation stage is pixel-level soft-label cross entropy loss to deal with both soft pseudo and fine-grained labels. 
During whole slide image classification stage, for each super size whole slide image, top $K$ patches with maximum pixel-level probability are the input to image classifier $\theta_{2}$. Average probability is the final image-level confidence to calculate loss with image-level labels. Stage2 training also decides the convergence criteria. We perform stage2 training for each round of stage1 and select that round whose stage2 training loss is the lowest. That means the top K patches selected in this round by stage1 is the optimal to fit image-level annotations in stage2.

\section{Experiments}

\subsection{Experimental setting}
To verify the effectiveness of hybrid supervision learning,we design three experiments of different supervision for comparison: \textbf{image-level}, \textbf{pixel-level} and \textbf{hybrid}. \textbf{Image-level} experiment is applied with source code from Campanella et al \cite{nature}, whose method relies only on image-level labels and searches top $K$ responsible patches for image-level labels. Patch classification model predicts every patch's confidence to be positive, then top $K$ patches are further trained by image-level labels.  It is $K$ = 1 in the released code \cite{nature}. \textbf{Pixel-level} experiment uses all of existing pixel-level fine-grained labels to train patch segmentation model and image-level labels to train whole slide image classification model, without extracting hidden pixel-level pseudo labels. We modified code from Khened et al \cite{khened2020generalized} as our pixel-level baseline. \textbf{Hybrid} experiment makes use of image-level labels, pixel-level fine-grained labels, and pseudo labels which are generated from the first two, as described above. We use normal classification task's evaluation metrics to evaluate the performance including Sensitivity, Specificity, ROC curve and AUC. Higher specificity while retaining 100\% sensitivity is our eventual preference.

\subsection{Implementation details}
Deep learning algorithms are developed by Pytorch1.0 \cite{[pytorch]} along with Openslide \cite{goode2013openslide}. Otsu's \cite{otsu1979threshold} method is performed to quickly extract valid patches that contain tissue. DLA34up and DLA34 \cite{DLA} are the patch segmentation and whole slide image classification models $\theta_{1}, \theta_{2}$ . Re-weighting constant is $V$ = 4.0 for all the experiments. This value only influences the training time, the lower $V$ is, the longer time it will take. $V$ bigger than 4.0 shows no significant improvement in speed. The pixel-wise maximum confidence patch selection threshold $T$ is 0.4. Higher threshold $T$ can exclude some positive images, thus degrades sensitivity. Lower threshold $T$ involves too much patches during training, leading to longer convergence time but nearly the same final performance. Patch size for both of patch segmentation and whole slide image classification is H = 512, W = 512, overlapped 128. For classification top $K$ patches, here we set K as 16. During patch segmentation model training, $R$, sampling ratio of pixel-level fine-grained labels $y_{p}$, pseudo labels $\hat{y_{p+}}$ and hard negative patches  $\hat{y_{p-}}$ is 2:1:7, to allow hard negative patches to overwhelm wrong patterns in noisy positive pseudo labels and keep pixel-level fine-grained labels participating in training procedure. For multi-round iteration, patch model initializes the weights from previous round, not from scratch. For whole slide image classifier, sampling ratio of positive and negative images is 1:1. Finally, the framework iterates to round 2 during experiments. During training each round, patch model is trained 30 epochs while whole slide image classifier is trained 15 epochs.

\begin{table}
\caption{Data distribution and train/test separation of gastric cancer dataset.}
\label{table}
\setlength{\tabcolsep}{3pt}
\begin{tabular}{llllll}
\hline\noalign{\smallskip}
 &  Pixel-level & Images  & Images & Total  &  Total\\
Distribution  &   Train  & Train  & Test  & Images & Patients\\
\noalign{\smallskip}
\hline
\noalign{\smallskip}
Positive &  200 patches & 585  & 499  &  1,084&  724 \\
Negative &  0  & 4,096  & 5,714  & 9,810 & 5941\\
Total & 200 patches & 4,681 & 6,213 & 10,894 & 6,665\\
\hline
\end{tabular}
\label{datasep}
\end{table}

\subsection{Dataset}
The dataset is especially developed for the commonly seen gastric cancer with 10,894 whole slide images in total. Data distribution of pixel-level fine-grained labeled patches, positive and negative whole slides, train/test separation are shown in Tab. \ref{datasep}, this ratio 1:9 of positive and negative is approximately the same distribution of clinical daily works. All the slides are automatically scanned by digital pathology scanner Leica Aperio AT2 at 20X magnification (0.50$\mu$m/pixel). The image-level annotation is either `Positive', which refers to low-grade intraepithelial neoplasia, high-grade intraepithelial neoplasia, adenocarcinoma, signet ring cell carcinoma, and poorly cohesive carcinoma, or `Negative', including chronic atrophic gastritis, chronic non-atrophic gastritis, intestinal metaplasia, gastric polyps, gastric mucosal erosion, etc.

\subsection{Results}

\begin{figure}[!t]
\centerline{\includegraphics[width=0.9\columnwidth]{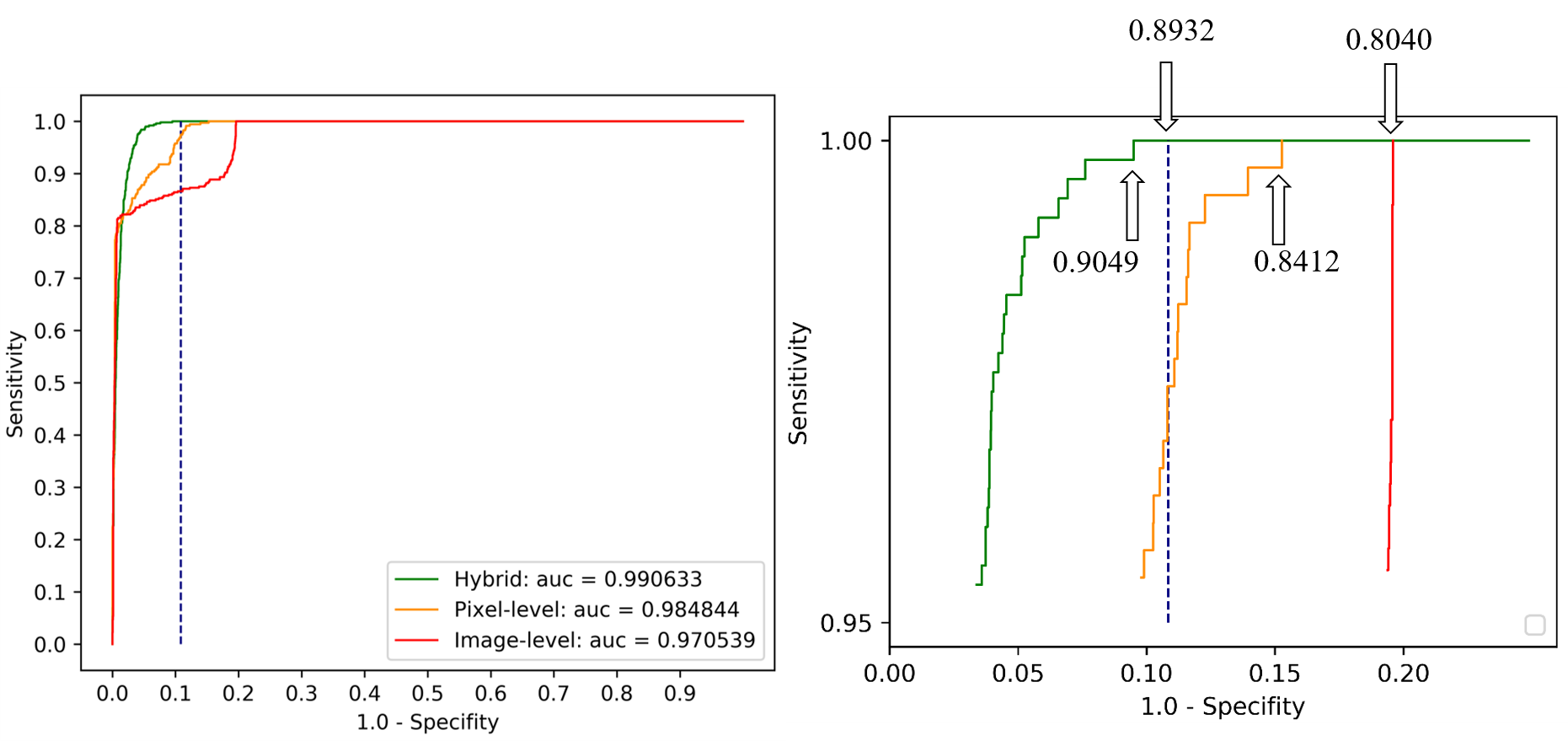}}
\caption{ROC curve detail of the three setting. In fact hybrid supervision learning can finally achieve specificity 0.9049 at probability threshold 0.1421.}
\label{auc}
\end{figure}

Released source code from Campanella et al. \cite{nature} running on our data is the compared image-level baseline for our hybrid supervision learning, which only uses the image-level label for training. This code can produce patch-level classification probability, not pixel-level. Statistically their weak supervision learning on our data achieve 80.40\% specificity while retaining 100\% sensitivity at threshold P = 0.0012. At the same time our hybrid supervision framework can achieve 89.32\% specificity with 100\% sensitivity at threshold P = 0.1000, 8.92\% far more than image-level annotation. These are summarized in Fig. \ref{auc}.

The reason for this phenomenon is that with image-level labels only, Campanella et al \cite{nature} searches the minimum representative patches responsible for image-level labels, without covering the most positive regions. At the same time our hybrid supervision learning framework is trained on most of positive patches in each positive whole slide image. More positive training samples establish more clear boundary between positive and negative patches, leading to higher specificity and area under curve.

\begin{table}
\caption{Statistics results of gastric cancer dataset.}
\label{table}
\setlength{\tabcolsep}{11pt}
\begin{tabular}{llll}
\hline\noalign{\smallskip}
Metrics  &   Hybrid  & Pixel-level \cite{khened2020generalized} & Image-level \cite{nature}\\
\noalign{\smallskip}
\hline
\noalign{\smallskip}
Sensitivity & 1.0000 &  1.0000  & 1.0000 \\
Specificity  &  0.8932  & 0.8412  & 0.8040\\
AUC & 0.9906 & 0.9848 & 0.9705 \\
Threshold & 0.1000 & 0.0041 & 0.0012 \\
\hline
\end{tabular}
\label{result1}
\end{table}

In pixel-level training, 200 pixel-level fine-grained labeled big patches are used to train the initial patch segmentation model then select top patches for whole slide images classifier training, without image-level labels guided pixel-level pseudo labels generation. This procedure can be regarded as backbone shared hybrid supervision learning, using existing annotations without extracting pixel-level pseudo labels in image-level positive images. This configuration provided us 84.12\% specificity, 5.2\% lower than hybrid supervision learning, indicating the proposed hybrid supervision learning is better than simple backbones sharing manner. As for thresholds, in test data hybrid supervision learning could reach 0.9049 specificity at threshold = 0.1421, while Pixel-level and Image-level have to use extremely low threshold for 100\% sensitivity, due to less supervision information.

\section{Conclusions}
In this paper we propose a novel hybrid supervision framework especially for whole slide images classification. Without entire image input to do end-to-end classification training. Pixel-level false negatives is prevented by re-weighting on pseudo labels of selected patches from positive images. False positives is suppressed by high training sampling ratio of hard negative patches. With this framework and a few pixel-level fine-grained labeled data, we can properly utilize large amount of image-level labeled whole slide images, train models in segmentation manner, and get much higher performance compared to using single format of annotation only, or just the existing pixel-level fine-grained labels.

%
%
%
 \bibliographystyle{splncs04}
 \bibliography{refs}

\end{document}